\documentclass[10pt,twocolumn,letterpaper]{article}

\usepackage{iccv}
\usepackage{times}
\usepackage{epsfig}
\usepackage{graphicx}
\usepackage{amsmath}
\usepackage{amssymb}

\newcommand{\pub}[1]{\color{gray}{\tiny{[{#1}]}}}
\usepackage{caption}
\usepackage{booktabs}
\usepackage{multirow}
\usepackage{pifont}
\usepackage{booktabs}
\usepackage{epsfig}
\usepackage{booktabs,multirow}
\usepackage{graphics}
\usepackage{threeparttable}
\usepackage{color}
\usepackage[normalem]{ulem}
\usepackage{multirow}

\definecolor{mygray}{gray}{.9}
\definecolor{gray}{RGB}{127,127,127}

\newcommand{\cmark}{\ding{51}}%
\newcommand{\xmark}{\ding{55}}%

\usepackage{circledtext}
\usepackage{pgffor}

\usepackage[breaklinks=true,bookmarks=false]{hyperref}

\iccvfinalcopy 

\def\iccvPaperID{****} 

\ificcvfinal\fi

\begin{document}

\title{\textsc{Kefa}: A Knowledge Enhanced and Fine-grained Aligned Speaker \\ for Navigation Instruction Generation}

\author{Haitian Zeng\\
University of Technology Sydney\\
{\tt\small haitian.zeng@student.uts.edu.au}
\and
Xiaohan Wang, Wenguan Wang, Yi Yang\\
Zhejiang University\\
{\tt\scriptsize wxh1996111@gmail.com, wenguanwang.ai@gmail.com, yangyics@zju.edu.cn}
}

\maketitle
\ificcvfinal\thispagestyle{empty}\fi

\begin{abstract}
   We introduce a novel speaker model \textsc{Kefa} for navigation instruction generation. The existing speaker models in Vision-and-Language Navigation suffer from the large domain gap of vision features between different environments and insufficient temporal grounding capability. To address the challenges, we propose a Knowledge Refinement Module to enhance the feature representation with external knowledge facts, and an Adaptive Temporal Alignment method to enforce fine-grained alignment between the generated instructions and the observation sequences. Moreover, we propose a new metric SPICE-D for navigation instruction evaluation, which is aware of the correctness of direction phrases. The experimental results on R2R and UrbanWalk datasets show that the proposed KEFA speaker achieves state-of-the-art instruction generation performance for both indoor and outdoor scenes.
\end{abstract}

\section{Introduction}

Vision-and-Language Navigation (VLN)~\cite{anderson2018vision} is a task where an agent follows natural language instructions to take actions and move to the destination in a virtual environment. While a great progress has been made in developing an instruction-following agent~\cite{ma2019regretful,wang2021structured,chen2022think}, the inverse task\textemdash \emph{instruction generation}, has received an increasing amount of attention recently. The instruction generation model, or a \emph{speaker}, usually plays the role of describing a trajectory in the environment with natural language. In practical scenarios, the speaker model can be used for describing a path explored by a robot to human in human-robot collaboration tasks~\cite{fong2003collaboration,st2015robot}, or guiding a blind follower with assistive instructions~\cite{huang2022assister}.



\begin{figure}[t]
\begin{center}
\includegraphics[width=0.9\linewidth]{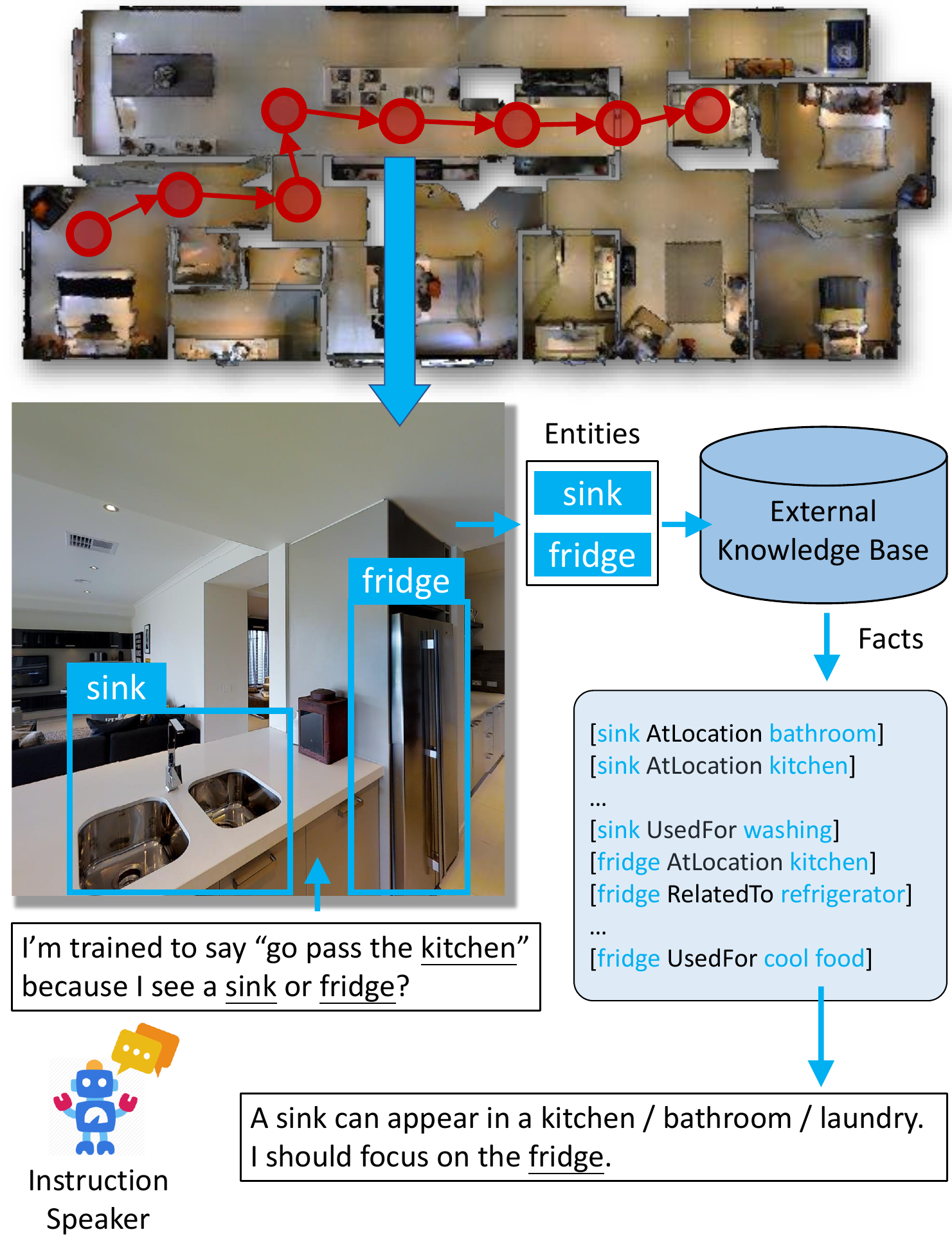}
\end{center}
\vspace{-1em}
\caption{Illustration. The speaker is strengthened with the commonsense that a sink can appear in several types of places but a fridge only appears in the kitchen.}
\vspace{-1em}
\label{fig:figure1}
\end{figure}

Despite the promising advances introduced by previous instruction generation approaches~\cite{fried2018speaker,agarwal2019visual,wang2022counterfactual,tan2019learning}, most of the speaker models still suffer from two major challenges. The first challenge is the significant difference of the visual observation features between the training and testing environment~\cite{zhang2020diagnosing}, which might reduce the generalization of the vision-based speaker model. The second challenge is the lack of temporal grounding capability of the speaker. The standard attention mechanism in the language decoder is not sufficient to capture the fine-grained correspondence between vision and language over time~\cite{ma2019self}, which might lead to hallucination in the generation.



To this end, we propose a novel \underline{K}nowledge \underline{E}nhance and \underline{F}ine-grained \underline{A}ligned (\textsc{Kefa}) speaker for instruction generation. We introduce two approaches, namely Knowledge Refinement Module (KRM) and Adaptive Temporal Alignment (ATA), to improve the performance of the speaker from two different aspects. First, to alleviate the undesirable domain gap of visual features, we leverage external knowledge to build a strong semantic clue for the language generation. An intuitive example is shown in Fig.~\!\ref{fig:figure1}. We humans have the commonsense knowledge that a place is most likely to be a kitchen because there is a fridge, while the speaker is unaware of such facts without training. To integrate such commonsense knowledge into the model, the Knowledge Refinement Module retrieves and encodes useful knowledge facts related to the observed objects in the environment from an external knowledge base like ConceptNet~\cite{speer2017conceptnet}. The encoded knowledge is then aggregated with the panorama features with a cross-modal attention mechanism. Second, to address the temporal grounding problem, the Adaptive Temporal Alignment method obtains the fine-grained correspondence between sub-instructions and viewpoints in a self-supervised way with the Dynamic Time Warping~\cite{berndt1994using} algorithm. Based on the fine-grained correspondence, we design an Attention Coverage Loss and a Contrastive Loss to enforce attention and feature alignment.


Moreover, we contribute a novel metric SPICE-D for the evaluation of navigation instruction generation. Previous metrics treat the directional phrases (\emph{e.g.} \texttt{turn} \texttt{right}) as a normal n-gram and do not emphasize the order of such phrases~\cite{papineni2002bleu,vedantam2015cider}. However, those direction phrases usually inform the instruction receiver of critical and sharp change of moving directions, which is important to the success of navigation. Motivated by this concern, we design the metric SPICE-D, which incorporates a new \emph{direction} component into the standard SPICE~\cite{anderson2016spice} score. The direction component is calculated based on a sequential matching of directional phrases of the candidate and reference instructions from the Longest Common Subsequence (LCS)~\cite{bergroth2000survey} algorithm. We show that SPICE-D is more correlated to the correctness of directions than the existing metrics.

In the experiment, we evaluate our \textsc{Kefa} speaker on the indoor R2R~\cite{anderson2018vision} dataset and the outdoor UrbanWalk~\cite{huang2022assister} dataset for instruction generation. Compared to the existing methods, the proposed \textsc{Kefa} speaker achieves state-of-the-art navigation instruction generation performance.


Our contributions are summarized as follows:

\begin{itemize}
    \item We propose a knowledge refinement module to leverage the semantic commonsense knowledge of detected objects. The refined feature improves the instruction generation capability of the model on unseen environments.

    \item We introduce an adaptive temporal alignment (ATA) method for improving the language grounding on a sub-instruction level. The ATA encourages the visual representation to be discriminative between the aligned and unaligned sub-instruction features by using an attention loss and a contrastive objective, while the alignment is adaptively obtained with Dynamic Temporal Warping. Benefiting from the ATA, the model shows an enhanced generation performance. 

    \item We introduce SPICE-D, a novel automatic metric for the evaluation of navigation instruction generation. SPICE-D emphasizes the matching of directional phrases, which is a critical component of navigation instructions, and it is not explicitly considered in the existing metrics. We show that SPICE-D is more correlated with the correctness of directional phrases than existing metrics.
\end{itemize}



\section{Related Works}

\noindent \textbf{Navigation Instruction Generation.} The earliest instruction generation dates back to~\cite{Lynch}, and it has been studied in various fields including robotics~\cite{goeddel2012dart}, cognitive science~\cite{kuipers1978modeling,evans1981environmental} and psychology~\cite{vanetti1988communicating}. Some early approaches~\cite{ward1986turn,allen1997knowledge,lovelace1999elements} are based on hand-crafted rules~\cite{dale2004using} and human design templates~\cite{look2005location} to produce route instructions from a map, following an easy-to-follow principle~\cite{goeddel2012dart}. Recent instruction generation methods progress with the development of Vision-and-Language Navigation (VLN). 
The speaker model is used for data augmentation and re-weighting the route choice for a navigation agent~\cite{fried2018speaker}. To train a better speaker, grounding to landmarks is emphasized in both seq-to-seq models~\cite{agarwal2019visual} and prompt-template based method~\cite{wang2022less}. Moreover, speakers are also integrated with the navigation agent in a counterfactual cycle-consistent learning~\cite{wang2022counterfactual} or a joint optimization~\cite{dou2022foam}, for their correlation can boost the performance.

\noindent \textbf{External Knowledge.} External knowledge from large Knowledge Bases like ConceptNet~\cite{speer2017conceptnet} and WordNet~\cite{fellbaum2010wordnet} is a rich source of commonsense facts, exploited in various tasks including visual question answering~\cite{shah2019kvqa,zhang2020rich,yu2020cross,luo2021weakly}, image captioning~\cite{wu2017image}, text generation~\cite{koncel2019text}, medication report generation~\cite{liu2021auto,li2019knowledge}, \emph{etc}. In VLN related tasks, Gao \emph{et al}.~\cite{gao2021room} use the external knowledge to reason over the room-and-object relations to boost the performance of agent on the remote referring expression task. Yang \emph{et al}.~\cite{yang2019visual} leverage the knowledge graph as scene prior for semantic navigation. Nevertheless, in navigation instruction generation, adopting external knowledge is still under-explored.

\noindent \textbf{Vision-Language Grounding.} Grounding language concepts to visual clues plays an important role in various visual recognition tasks, including image captioning~\cite{zhou2020more}, visual question answering~\cite{luo2021weakly}, language-image pre-training~\cite{li2020oscar}, text-video grounding~\cite{chen2022explore}, phrase and referring expression grounding~\cite{huang2018finding,liu2019knowledge}, temporal action localization~\cite{luo2021action}, \emph{etc}. For an instruction following agent, Hu \emph{et al}.~\cite{hu2019you} propose to ground language instructions to visual and geometric modalities using a mixture of experts approach. VLN-BERT~\cite{hong2020recurrent} uses a large amount image-text pairs from the web and finetunes a ViLBERT~\cite{lu2019vilbert} to model the trajectory-instruction compatibility. Despite these advances, the vision-language grounding is less explored in a sequence-to-sequence instruction generator.


\noindent \textbf{Evaluation Metrics of Language Generation.} 
There are a number of automatic evaluation metrics for text similarity evaluation. In machine translation and image captioning, BLEU~\cite{papineni2002bleu}, METEOR~\cite{banerjee2005meteor}, ROUGE~\cite{lin2004rouge} CIDEr~\cite{vedantam2015cider} and SPICE~\cite{anderson2016spice} are most frequently adopted, and they are also used for navigation instruction evaluation. With the emerging pre-trained large language models like BERT~\cite{kenton2019bert}, model-based metrics like BERTScore~\cite{zhangbertscore} are introduced. However, these metrics have placed less emphasis on the directional phrases, which are important components of navigation instructions and have to appear in a correct order.



\begin{figure*}[t]
\begin{center}
\includegraphics[width=1.0\linewidth]{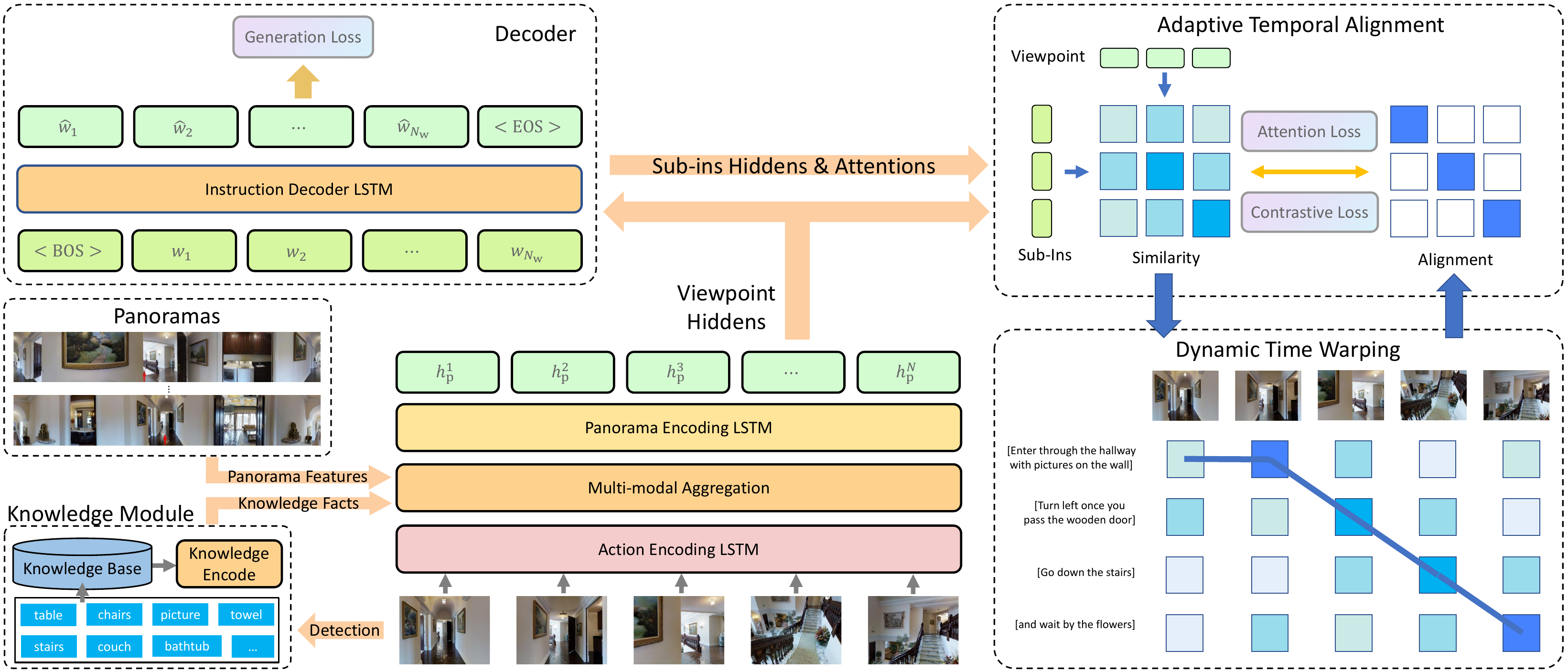}
\end{center}
\vspace{-1em}
\caption{Overview. In the encoder, \textsc{Kefa} speaker fuses the semantic knowledge from the Knowledge Base ~\cite{speer2017conceptnet} with the panorama vision features (\S\ref{sec:encoder}). In decoding, \textsc{Kefa} is supervised with an Adaptive Temporal Alignment (ATA) method. In ATA, the similarity of the hidden states of viewpoints and sub-instructions are fed into the Dynamic Time Warping algorithm to obtain an temporal alignment, which is further used to calculate the Attention Coverage and Contrastive Loss. (\S\ref{sec:decoder})}
\vspace{-1em}
\label{fig:figure2}
\end{figure*}

\section{Methods}

\subsection{Task Formulation}
Given a navigation trajectory, the task of navigation instruction generation is to produce an instruction $\boldsymbol{I}=(w_1, \cdots, w_{N_\mathrm{w}})$ that can guides human or virtual agents to travel along the trajectory in an unfamiliar environment. The trajectory is a sequence of visual observations $\boldsymbol{T}=(\boldsymbol{v}_1, \cdots, \boldsymbol{v}_N)$. The objective is to learn a model with parameter $\boldsymbol{\theta}$ by maximizing the likelihood of the target instruction $\boldsymbol{I}$ conditioned on the input trajectory $\boldsymbol{T}$:
\begin{eqnarray}
\boldsymbol{\theta}^* = \mathop{\arg \min_{\boldsymbol{\theta}}} \log p(\boldsymbol{I} \mid \boldsymbol{T};\boldsymbol{\theta}).
\label{eq:objective}
\end{eqnarray}
To realize the generation, the model generally consists of a trajectory encoder $\mathcal{E}$  and a language decoder $\mathcal{D}$. We describe the encoder in \S\ref{sec:encoder} and the decoder in \S\ref{sec:decoder}. 

\subsection{Knowledge Refined Trajectory Encoding}
\label{sec:encoder}
We introduce the encoder $\mathcal{E}$, which encodes the visual observations and enhance the feature with common-sense knowledge. At step $t$ in the trajectory, the visual observation $\boldsymbol{v}_t$ is a panorama consisting of $N_\mathrm{v}$ views, which can be written as: $\boldsymbol{v}_t=(v_{t,1}, \cdots, v_{t,N_\mathrm{v}})$. Consistent with previous works~\cite{fried2018speaker,agarwal2019visual,tan2019learning,wang2022counterfactual}, we discretize the 360-degree panoramic view into 12 horizontal angles and 3 vertical elevations, resulting in a total of $N_\mathrm{v}=36$ views. At each step, the view that is most close to the direction of the next step in the trajectory is the action view $v_{t,a}, a \in \left[ 1, N_\mathrm{v} \right]$.

To encode the visual features of the trajectory, the images of all views are first processed with a backbone network. We adopt ResNet-152~\cite{he2016deep} pre-trained on ImageNet~\cite{ILSVRC15} as the backbone $\mathcal{F}$ to extract image features of each view. The backbone network is frozen and will not receive gradients in the training. Next, the features of the action views in the trajectory are encoded with a LSTM~\cite{hochreiter1997long}:
\begin{eqnarray}
h^t_\mathrm{a} = \mathrm{LSTM_a}(\mathcal{F}(v_{t,a}), h^{t-1}_\mathrm{a}), \quad t=1 \cdots N,
\label{eq:objective}
\end{eqnarray}
where $h^t_\mathrm{a}$ is the hidden state of the LSTM at time step $t$.

\noindent \textbf{Environment Entity Acquiring with Detection.} We start the knowledge enhancement with the detection from the action views. The objects in the action views are most relevant to the target instruction. Therefore, for each action view $v_{t,a}$ in the trajectory, we extract detection results using an off-the-shelf object detector Faster-RCNN~\cite{ren2015faster} trained on the visual genome~\cite{krishna2017visual} dataset. The detector outputs the detection results with confidence scores in 1600 object categories, which captures most of the diversity of the landmark objects appearing the navigation environment. We filter the detection results with confidence larger than 0.5 and gather the class tags of the detections as an entity set $\boldsymbol{E}_t=\{E_{t,1}, \cdots, E_{t,N_E}\}$ for $\boldsymbol{v}_t$.

\noindent \textbf{Knowledge Retrieval and Embedding.} In this step, we aims to enhance the action sequence representation with useful external knowledge. The external knowledge brings extra common-sense relationships to the instruction generation process. For instance, if a \texttt{Microwave} appears in the entity set $\boldsymbol{E}_t$, the knowledge fact \texttt{Microwave} \texttt{AtLocation} \texttt{Kitchen} will give more evidence to generator when it is required to output an instruction like \texttt{Go} \texttt{into} \texttt{the} \texttt{Kitchen}. To achieve these, we match each element $E_{t,i}$ of $\boldsymbol{E}_t$ to the semantic entities in the external Knowledge Base (\emph{i.e.} Concept-Net~\cite{speer2017conceptnet}) and retrieve the top-$K$ corresponding knowledge facts:
\begin{eqnarray}
\boldsymbol{f}_{t,i} = \left \{ \left \langle e_{t,i}^{j}, rel_{t,i}^{j}, z_{t,i}^{j} \right \rangle \right \} \quad j=1\cdots K,
\end{eqnarray}
where $e_{t,i}^{j}$ is the related entity of the knowledge fact, $rel_{t,i}^{j}$ is the relation type (\emph{e.g.} \texttt{AtLocation}, \texttt{RelatedTo}), $z_{t,i}^{j}$ is the weight of relation used for ranking the top-$K$ facts. To encode the knowledge facts, we first embed each $e_{t,i}^{j}$ and $rel_{t,i}^{j}$ with GloVe~\cite{jeffreypennington2014glove} embedding, then apply a knowledge summary module $\mathcal{G}$ which consists of a convolution layer followed by an average-pooling layer:
\begin{eqnarray}
\mathcal{G}(\boldsymbol{f}_{t,\cdot}) = \mathrm{AvgPool}\circ\mathrm{Conv} \left(\left[ W_\mathrm{G}e_{t,i}^{j} \parallel W_\mathrm{G}rel_{t,i}^{j}\right]\right),
\end{eqnarray}
where $W_\mathrm{G}$ is the GloVe embedding matrix, $\left[ \cdot \right]$ denotes the concatenate operation, the average-pooling is conducted over all $i,j$.
We empirically found this design is more robust than other choices like GRU~\cite{chung2014empirical} or self-attention~\cite{vaswani2017attention} encoding which may cause over-fitting of the module. Finally, we denote the encoded knowledge vector of each action view as $\mathcal{G}(\boldsymbol{f}_{t,\cdot}) =know_{t} \in \mathbb{R}^{D_k}$.

\noindent \textbf{Panoramic Multi-modal Aggregation.} We now introduce a new mechanism to aggregate the external knowledge with the panorama views. With the entity set $\boldsymbol{E}_t$, we first create a multi-modal feature of the panorama views:
\begin{eqnarray}
v_{t,l}^+=\left[ \mathcal{F}(v_{t,l}) \parallel \bar{E}_t \right] \quad l = 1\cdots N_\mathrm{v},
\end{eqnarray}
where $\bar{E}_t$ is the mean GloVe embedding of detected entities. Then we design an attention mechanism to aggregate the panorama views together with the knowledge vector $know_{t}$, written as:
\begin{align*}
& \alpha_l=\frac{(W_\mathrm{q} h^t_\mathrm{a})^T(W_\mathrm{k} v_{t,l}^+)}{\Sigma_{l^{\prime}}(W_\mathrm{q} h^t_\mathrm{a})^T(W_\mathrm{k} v_{t,l^{\prime}}^+)}, \\
& g_t=\tanh(W_\mathrm{out}(\left[ \Sigma_{l^{\prime}} (\alpha_{l^{\prime}} \mathcal{F}(v_{t,l^{\prime}})) \parallel h^t_\mathrm{a} \parallel know_{t} \right])),
\end{align*}
where $W_\mathrm{q},W_\mathrm{k},W_\mathrm{out}$ are query, key and output linear layers that map the corresponding inputs into dimension $D_h$.

Following~\cite{tan2019learning}, we use another LSTM layer to model the sequence of aggregated multi-modal panorama features $h^t_\mathrm{p}$:
\begin{eqnarray}
h^t_\mathrm{p} = \mathrm{LSTM_p}(g_t, h^{t-1}_\mathrm{p}), \quad t=1\cdots N.
\label{eq:objective}
\end{eqnarray}

\subsection{Adaptive Temporal Alignment Guided Decoding}
\label{sec:decoder}
In this part, we introduce a novel Adaptive Temporal Alignment (ATA) method for improving the sub-instruction level language grounding of the decoder $\mathcal{D}$. For a given trajectory, the navigation instruction usually consists of multiple actions, \emph{e.g.} \texttt{Walk} \texttt{out} \texttt{of} \texttt{the} \texttt{bathroom} \texttt{and} \texttt{go} \texttt{into} \texttt{the} \texttt{living} \texttt{room}. This indicates that there exists a fine-grained correspondence between the instruction and trajectory. However, this correspondence cannot be successfully captured by the vanilla attention decoder~\cite{ma2019self}, and it is also less considered in an instruction generator. 

Motivated by the above concern, we develop an attention alignment loss and a contrastive loss for strengthening the awareness of the fine-grained correspondence in the decoding stage. We hypothesize that the awareness of such correspondence is beneficial for the generator to produce more grounded navigation instructions and better generalize on unseen trajectories. To realize such guidance to the decoder, the alignment between sub-instructions and sub-trajectories must be found first. To calculate the alignment, we first chunk the instructions into a sequence of sub-instructions by a language dependency parser, and then match them to sub-trajectories. However, the major difficulty encountered here is that the partition of sub-trajectories is hard to obtain.

Different from~\cite{hong2020sub} where the sub-trajectory partition is manually annotated by a crowd-sourcing platform, we propose a novel \emph{self-supervised} method to obtain the alignment between sub-instructions and sub-trajectories. We leverage the most important property \emph{monotonicity} of the correspondence. Given the hidden states of panorama features and decoded instructions, we adaptively obtain the temporal alignment by using the Dynamic Time Warping (DTW)~\cite{berndt1994using} algorithm, and use the alignment as the signal for learning better attention coverage and discriminative features.

 Specifically, suppose that the navigation instruction $\boldsymbol{I}$ can be broken down into a sequence of sub-instructions: $\boldsymbol{I}=(\boldsymbol{I}_{\mathrm{sub}}^{1}, \cdots, \boldsymbol{I}_{\mathrm{sub}}^{M})$, and each sub-instruction is generally an action phrase like \texttt{go} \texttt{into} \texttt{the} \texttt{living} \texttt{room}. To achieve the sub-instruction parsing, we follow~\cite{hong2020sub} to chunk the original instruction with the dependency parser from the Stanford NLP library~\cite{manning2014stanford}. The dependency parser assigns each word in the sentence to a dependency tree where the root is an action word. Next, we denote the mean hidden states of the words of $m$-th sub-instruction as $h^m_\mathrm{I}$. We feed the sequences of panorama and sub-instructions hidden states to the Dynamic Time Warping~\cite{berndt1994using} algorithm to obtain an alignment:
\begin{eqnarray}
\boldsymbol{A} = \mathrm{DTW}\left( \left \{ h^m_\mathrm{I} \right \}_{m=1}^M, \left \{ h^t_\mathrm{p} \right \}_{t=1}^N \right),
\end{eqnarray}
where $\boldsymbol{A} \in \mathbb{R}^{M \times N}$ is the obtained alignment matrix and the $(m, t)$ element of $\boldsymbol{A}$ is:
\begin{eqnarray}
A_{m, t} =  
\left\{
\begin{aligned}
1 & , & \text{$\boldsymbol{I}_{\mathrm{sub}}^{m}$ is aligned to viewpoint $t$}, \\
0 & , & \text{otherwise}.
\end{aligned}
\right.
\end{eqnarray}
In other words, the alignment matrix represents whether a sub-instruction is aligned to a viewpoint in the trajectory. With the alignment matrix $\boldsymbol{A}$, we design the following two losses to enforce the temporal alignment:

\noindent \textbf{Attention Coverage Loss.} This loss function encourages the attention weights of each generated word to concentrate on the aligned sub-trajectories. In each step of instruction decoding, the decoder outputs a word $w_o$ by attending to the panorama feature $h_\mathrm{p}^t$ with the attention weights $\beta_{o,t}$. We calculate a target matrix $\boldsymbol{A}^{\prime} \in \mathbb{R}^{N_\mathrm{w} \times N}$ based on $\boldsymbol{A}$:
\begin{eqnarray}
A^{\prime}_{o, t} =  
\left\{
\begin{aligned}
1 & , & w_o \in \boldsymbol{I}_{\mathrm{sub}}^{m} \text{ and } A_{m, t}=1, \\
0 & , & \text{otherwise}.
\end{aligned}
\right.
\end{eqnarray}
Now we use $\boldsymbol{A}^{\prime}$ as the target to compute the attention coverage loss, which is:
\begin{eqnarray}
\begin{split}
\mathcal{L}_{\mathrm{att}} = -\frac{1}{N_\mathrm{w}}\sum\limits_{o=1}^{N_\mathrm{w}}\left( \log \left( \sum\limits_{t=1}^{N} A^{\prime}_{o, t}\beta_{o,t} \right) \right. \\
\left. + \log \left( \sum\limits_{t=1}^{N} \left(1-A^{\prime}_{o, t}\right)\left(1-\beta_{o,t} \right)\right) \right).
\end{split}
\end{eqnarray}
Intuitively, this loss sums up the attention weights on the aligned viewpoints and unaligned viewpoints respectively, and then calculates the score using binary cross entropy.

\noindent \textbf{Contrastive Loss.} This objective enforces panorama features and language hidden states to be discriminative between different sub-instructions and sub-trajectories. We denotes the hidden state of $w_o$ in decoder as $h^o_{\mathrm{lan}}$. The contrastive loss is computed by:
\begin{eqnarray}
\mathcal{L}_{\mathrm{nce}} = -\frac{1}{N_\mathrm{w}}\sum\limits_{o=1}^{N_\mathrm{w}}\left(
\log\frac{\sum\limits_{t=1}^{N}{A^{\prime}_{o, t}\mathrm{exp}(h_\mathrm{p}^t \cdot h^o_{\mathrm{lan}}})}{\sum\limits_{t=1}^{N}{\mathrm{exp}(h_\mathrm{p}^t \cdot h^o_{\mathrm{lan}}})} \right).
\end{eqnarray}

\noindent \textbf{Total Objective.} 
The final training loss of the model is the sum of three losses:
\begin{eqnarray}
\mathcal{L} = \mathcal{L}_{\mathrm{CE}}(\boldsymbol{I}, \boldsymbol{\hat{I}}) + \lambda_1\mathcal{L}_{\mathrm{att}} + \lambda_2\mathcal{L}_{\mathrm{nce}},
\end{eqnarray}
where $\lambda_1$ and $\lambda_2$ are balancing factors, $\mathcal{L}_{\mathrm{CE}}$ is the cross entropy loss, $\boldsymbol{\hat{I}}=\mathcal{D}\left(\mathcal{E}(\boldsymbol{T})\right)$ is the generated instruction.


\section{SPICE-D: Direction-Aware Evaluation of Navigation Instructions}
\label{sec:spice_d}

In this section, we introduce a novel automatic evaluation metric SPICE-D for assessing the quality of the navigation instructions. We first review the SPICE~\cite{anderson2016spice} metric and then introduce the new metric. 

\subsection{SPICE Recap}
Semantic Propositional Image Caption Evaluation (SPICE)~\cite{anderson2016spice} is a metric for the evaluation of image captioning. SPICE focuses on the conceptual correctness of the caption by using a scene graph representation~\cite{johnson2015image}. The scene graph contains three types of key components as the vertices: objects, relations, and attributes, and the vertices are connected by corresponding edges. SPICE parses the caption into scene graphs using the off-the-shelf parser, and then extracts all tuples from the graph. The tuples from candidate (prediction) and reference (ground-truth) are marked as $\mathrm{\mathbf{F}^{C}}$ and $\mathrm{\mathbf{F}^{R}}$, respectively. Next, the precision ($\mathrm{Pr}_{\,\mathrm{S}}$) and recall ($\mathrm{Re}_{\,\mathrm{S}}$) of SPICE are calculated as:
\begin{eqnarray}
\begin{aligned}
\mathrm{Pr}_{\,\mathrm{S}} = \frac{\left|\mathrm{\mathbf{F}^{C}}\cap \mathrm{\mathbf{F}^{R}}\right|}{\left| \mathrm{\mathbf{F}^{C}}\right|}, \quad
\mathrm{Re}_{\,\mathrm{S}} = \frac{\left|\mathrm{\mathbf{F}^{C}}\cap \mathrm{\mathbf{F}^{R}}\right|}{\left| \mathrm{\mathbf{F}^{R}}\right|}.
\end{aligned}
\label{eq:pr_re}
\end{eqnarray}
The SPICE is the F-score of the precision and recall:
\begin{eqnarray}
\mathrm{SPICE} =
\frac{2}{1/\mathrm{Pr}_{\,\mathrm{S}} + 1/\mathrm{Re}_{\,\mathrm{S}}}.
\end{eqnarray}

%

\begin{figure*}[t]
\begin{center}
\includegraphics[width=1.0\linewidth]{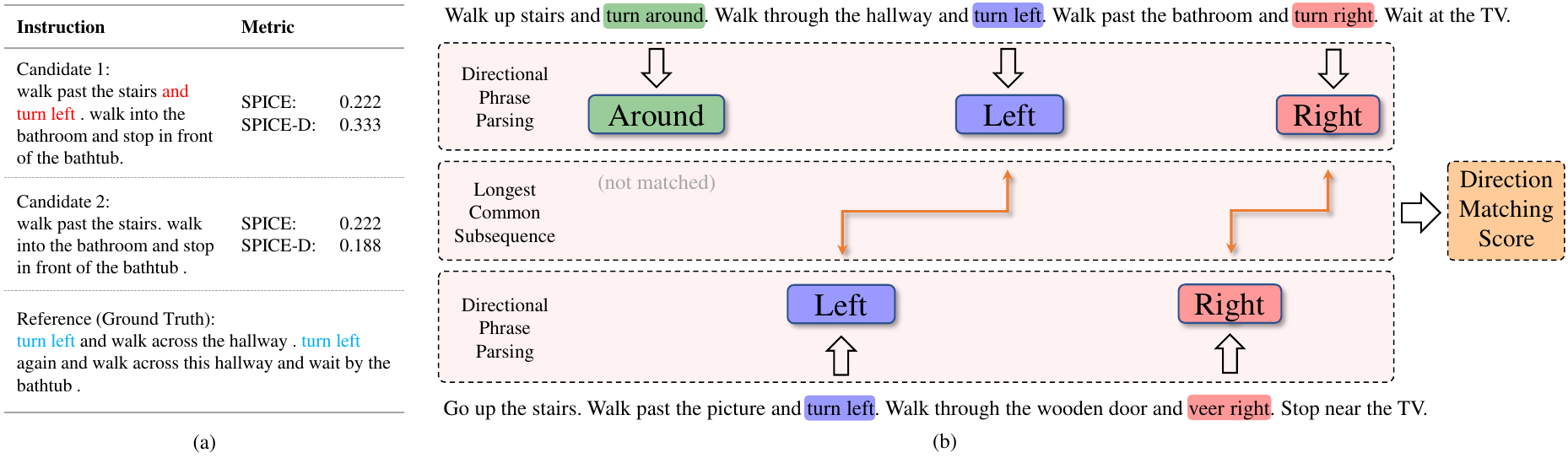}
\end{center}
\vspace{-1em}
\caption{(a) An example of the SPICE~\cite{anderson2016spice} and our SPICE-D metric. The directional phrase \texttt{and} \texttt{turn} \texttt{left} does not affect the SPICE score in this example, while the SPICE-D score is sensitive to the difference. (b) The calculation of SPICE-D. Directional phrases are first parsed from the instructions and then matched with Longest Common Subsequence. (\S\ref{sec:spice_d_calculate})}
\vspace{-1em}
\label{fig:spice_d}
\end{figure*}

\subsection{SPICE-D Metric}
\label{sec:spice_d_calculate}
The current language evaluation metrics, including Bleu~\cite{papineni2002bleu}, Meteor~\cite{banerjee2005meteor}, Rouge~\cite{lin2004rouge} CIDEr~\cite{vedantam2015cider} and SPICE~\cite{anderson2016spice}, are developed mainly for evaluating the machine translation or image captioning tasks, and they are also widely-adopted in the evaluation navigation instruction generation. Among them, SPICE is considered to be most suitable for navigation instruction evaluation~\cite{zhao2021evaluation}. Despite their success, the directional pharses (\emph{e.g.} \texttt{turn} \texttt{right}) are not emphasized in the current metrics. For example, Fig.~\ref{fig:spice_d}(a) shows a case where SPICE is less sensitive to the correctness of directional phrases. This characteristic is undesirable in assessing the quality of generated instructions, for reason that these directional phrases usually mention critical and sharp changes of moving direction in the navigation process.

With this motivation, we introduce a novel SPICE-D metric, which is developed to meet the requirement of assessing the directional phrases together with the existing aspects of language quality. We introduce the directional phrase matching quality as a new component along with the original components in SPICE and takes it into the computation of the overall F-score as the SPICE-D score.

The calculation process of SPICE-D is shown in Fig.~\ref{fig:spice_d}(b). Given a candidate instruction generated by the model and the ground truth instruction, we first parse the directional phrases $\mathrm{\mathbf{D}^{C}}=(\mathrm{D_1^{C}},\cdots,\mathrm{D_{N_C}^{C}})$ from the candidate instruction and $\mathrm{\mathbf{D}^{R}}=(\mathrm{D_1^{R}},\cdots,\mathrm{D_{{N_R}}^{R}})$ from the ground truth instruction, where the number of phrases are $\mathrm{N_C}$ and $\mathrm{N_R}$, respectively. To handle the diverse expression of directional phrases, we create synonyms sets for each direction and group the phrases according to their directions. For example, \texttt{turn} \texttt{right}, \texttt{make} \texttt{a} \texttt{right} and \texttt{veer} \texttt{right} are all classified as the turning-right phrases. 

To measure the quality of directional phrases alignments, we use the Longest Common Subsequence (LCS)~\cite{bergroth2000survey} algorithm to obtain a mapping between $\mathrm{\mathbf{D}^{C}}$ and $\mathrm{\mathbf{D}^{R}}$. The directional phrases belong to the same class are considered as matched phrases. The length of the LCS is viewed as the number of successfully matched phrases, which is:
\begin{eqnarray}
  \left|\mathrm{\mathbf{D}^{C}}\cap \mathrm{\mathbf{D}^{R}}\right| = 
  \left|\mathop{\mathrm{LCS}}	\left(\mathrm{\mathbf{D}^{C}}, \mathrm{\mathbf{D}^{R}}	\right)\right|.
\label{eq:len_lcs}
\end{eqnarray}
The LCS algorithm ensures the directional phrases in the candidate and the reference to be matched uniquely and sequentially. Next, the precision and recall are computed by:
\begin{eqnarray}
\begin{aligned}
\mathrm{Pr}_{\,\mathrm{SD}} = \frac{\left|\mathrm{\mathbf{F}^{C}}\cap \mathrm{\mathbf{F}^{R}}\right| + \left|\mathrm{\mathbf{D}^{C}}\cap \mathrm{\mathbf{D}^{R}}\right|}{\left| \mathrm{\mathbf{F}^{C}}\right| + \left| \mathrm{\mathbf{D}^{C}}\right|}, \\
\mathrm{Re}_{\,\mathrm{SD}} =
\frac{\left|\mathrm{\mathbf{F}^{C}}\cap \mathrm{\mathbf{F}^{R}}\right| + \left|\mathrm{\mathbf{D}^{C}}\cap \mathrm{\mathbf{D}^{R}}\right|}{\left| \mathrm{\mathbf{F}^{R}}\right| + \left| \mathrm{\mathbf{D}^{R}}\right|}.
\end{aligned}
\label{eq:pr_re}
\end{eqnarray}
Finally, the SPICE-D score is the F-score based on the direction-calibrated precision and recall, which is:
\begin{eqnarray}
\mathrm{SPICE\text{-}D} =
\frac{2}{1/\mathrm{Pr}_{\,\mathrm{SD}} + 1/\mathrm{Re}_{\,\mathrm{SD}}}.
\end{eqnarray}

\begin{table*}[!bth]
	\centering
			\resizebox{0.8\textwidth}{!}{
			\setlength\tabcolsep{3pt}
			\renewcommand\arraystretch{1.0}
	\begin{tabular}{|rl||ccccccc|}
	\hline
	~ & & \multicolumn{7}{c|}{R2R \texttt{val} \texttt{seen}}\\
	\cline{3-9}\cline{3-9}
	\multicolumn{2}{|c||}{\multirow{-2}{*}{Methods}}
	&\textbf{\texttt{SPICE}}\!~$\uparrow$ &\textbf{\texttt{SPICE-D}}\!~$\uparrow$ &\texttt{Bleu-1}\!~$\uparrow$ &\texttt{Bleu-4}\!~$\uparrow$ &\texttt{CIDEr}\!~$\uparrow$ &\texttt{Meteor}\!~$\uparrow$ &\texttt{Rouge}\!~$\uparrow$ \\
	\hline
	\hline
	BT-speaker~\cite{fried2018speaker}&\!\!\pub{NeurIPS2018}    &22.1 &28.2 &72.2 &28.3 &43.7 &23.0 &49.5\\
        LandmarkSelect~\cite{agarwal2019visual}&\!\!\pub{CVPRW2019}    &21.4 & - &54.9 &15.7 &13.7 &22.8 &35.2\\
        EnvDrop~\cite{tan2019learning}&\!\!\pub{NAACL2019}    &24.3 &30.5 &72.5 &27.7 &47.8 &24.5 &49.6\\
        CCC~\cite{wang2022counterfactual}&\!\!\pub{CVPR2022}    &23.1 &29.9 &72.8 &28.7 &\textbf{54.3} &23.6 &49.3\\
\hline
        \textsc{Kefa} Speaker~&\!\!\footnotesize{(\textbf{\texttt{ours}})}    &\textbf{27.2} &\textbf{32.5} &\textbf{76.7} &\textbf{32.6} &52.3 &\textbf{25.4} &\textbf{51.8}\\
\hline
\hline
~ & & \multicolumn{7}{c|}{R2R \texttt{val} \texttt{unseen}}\\
	\cline{3-9}\cline{3-9}
	\multicolumn{2}{|c||}{\multirow{-2}{*}{Methods}}
	&\textbf{\texttt{SPICE}}\!~$\uparrow$ &\textbf{\texttt{SPICE-D}}\!~$\uparrow$ &\texttt{Bleu-1}\!~$\uparrow$ &\texttt{Bleu-4}\!~$\uparrow$ &\texttt{CIDEr}\!~$\uparrow$ &\texttt{Meteor}\!~$\uparrow$ &\texttt{Rouge}\!~$\uparrow$ \\
\hline
	BT-speaker~\cite{fried2018speaker}&\!\!\pub{NeurIPS2018}    &18.9 &25.1 &68.2 &26.3 &37.9 &21.7 &48.0\\
        LandmarkSelect~\cite{agarwal2019visual}&\!\!\pub{CVPRW2019}    &19.7 & - &54.8 &15.9 &13.2 &23.1 &35.7\\
        EnvDrop~\cite{tan2019learning}&\!\!\pub{NAACL2019}    &21.8 &28.0 &72.3 &27.1 &41.7 &23.6 &49.0\\
        CCC~\cite{wang2022counterfactual}&\!\!\pub{CVPR2022}    &21.4 &27.8 &70.8 &27.2 &\textbf{46.1} &23.1 &47.7\\
\hline
        \textsc{Kefa} Speaker~&\!\!\footnotesize{(\textbf{\texttt{ours}})}    &\textbf{23.4} &\textbf{29.3} &\textbf{73.8} &\textbf{28.3} &42.7 &\textbf{24.4} &\textbf{50.3}\\
\hline

	\end{tabular}
	}
		\vspace*{-2pt}
	\captionsetup{font=small}
		\caption{\small{Quantitative comparison results (\S\ref{sec:main_results}) on R2R~\cite{anderson2018vision} dataset.}}
		\label{table:r2r_results}
\end{table*}

\begin{table*}[!bth]
	\centering
			\resizebox{0.8\textwidth}{!}{
			\setlength\tabcolsep{3pt}
			\renewcommand\arraystretch{1.0}
	\begin{tabular}{|rl||ccccccc|}
	\hline
	~ & & \multicolumn{7}{c|}{UrbanWalk}\\
	\cline{3-9}\cline{3-9}
	\multicolumn{2}{|c||}{\multirow{-2}{*}{Methods}}
	&\textbf{\texttt{SPICE}}\!~$\uparrow$ &\textbf{\texttt{SPICE-D}}\!~$\uparrow$ &\texttt{Bleu-1}\!~$\uparrow$ &\texttt{Bleu-4}\!~$\uparrow$ &\texttt{CIDEr}\!~$\uparrow$ &\texttt{Meteor}\!~$\uparrow$ &\texttt{Rouge}\!~$\uparrow$ \\
	\hline
	\hline
	BT-speaker~\cite{fried2018speaker}&\!\!\pub{NeurIPS2018}    &52.4 &55.5 &64.9 &40.8 &92.8 &35.0 &62.0\\
        EnvDrop~\cite{tan2019learning}&\!\!\pub{NAACL2019}    &53.1 &56.0 &68.9 &43.5 &131.8 &35.8 &63.4\\
        ASSISTER*~\cite{huang2022assister}&\!\!\pub{ECCV2022}    &45.1 &47.9 &57.6 &16.4 &32.2 &31.9 &55.7\\
\hline
        \textsc{Kefa} Speaker~&\!\!\footnotesize{(\textbf{\texttt{ours}})}    &\textbf{56.6} &\textbf{59.2} &\textbf{71.1} &\textbf{45.0} &\textbf{139.9} &\textbf{37.8} &\textbf{65.5}\\
\hline
	\end{tabular}
	}
		\vspace*{-2pt}
	\captionsetup{font=small}
		\caption{\small{Quantitative comparison results (\S\ref{sec:main_results}) on UrbanWalk~\cite{huang2022assister} dataset.} * denotes our implementation. }
		\label{table:results_urbanwalk}
	\vspace*{-12pt}
\end{table*}

\section{Experiments}

\subsection{Experimental Setup}
\noindent \textbf{Datasets.} We evaluate our method on both indoor and outdoor navigation instruction generation datasets.

\textbf{R2R.} The Room-to-Room (R2R) Navigation dataset~\cite{anderson2018vision} is created under the Matterport3D environment. It contains a total of 7,189 paths in the environment and three human annotated navigation instructions for each path. The average length of the paths is 10 meters and the average length of the instructions is approximately 29 words. The dataset is split into \texttt{train}, \texttt{val} \texttt{seen} and \texttt{val} \texttt{unseen} which contains 14025, 1020 and 2349 instructions, respectively.

\textbf{UrbanWalk.} UrbanWalk dataset~\cite{huang2022assister} is a recent dataset for evaluating navigation instruction generation in outdoor scene. Due the dataset availability, we use the 26808 images from the simulation benchmark of UrbanWalk and sampled a total of 5349 routes from them. Each route consists of 5 to 7 viewpoints, and the instruction has around 100 words.

\noindent \textbf{Metric.} To evaluate the generated instructions, we report both the standard language metrics and the proposed SPICE-D metric. For the standard metrics, we follow previous work~\cite{fried2018speaker, tan2019learning, wang2022counterfactual} to use Bleu~\cite{papineni2002bleu}, Meteor~\cite{banerjee2005meteor}, Rouge~\cite{lin2004rouge}, CIDEr~\cite{vedantam2015cider} and SPICE~\cite{anderson2016spice}.

\subsection{Implementation Details}

We implement our method with PyTorch~\cite{paszke2019pytorch} library. The instruction speaker is trained with Adam~\cite{kingma2015adam} Optimizer for 80k iterations. Consistent to previous works~\cite{fried2018speaker,tan2019learning,agarwal2019visual,wang2022counterfactual}, the action view and panoramic visual features are extracted from ResNet-152~\cite{he2016deep} model and an angle feature of 128 size is concatenated. We adopt the ConceptNet~\cite{speer2017conceptnet} and retrieve $K=3$ commonsense knowledge facts. The dimension of hidden state $D_h$ is set to 512, and the knowledge vector dimension $D_k$ is 100. For the SPICE-D metric, we divide the directions into three categories \texttt{right},\texttt{left},\texttt{around} on R2R dataset. For the outdoor directions on UrbanWalk, the directions are classified into \texttt{right}, \texttt{left}, and \texttt{nine} to \texttt{three} o'clock. 

\noindent \textbf{Competitors.}
We compare our method with state-of-the-art navigation instruction generation methods, include speaker-follower~\cite{fried2018speaker}, LandmarkSelect~\cite{agarwal2019visual}, EnvDrop~\cite{tan2019learning} and CCC~\cite{wang2022counterfactual}. These methods all adopt ResNet-152 as the backbone network to extract the panorama visual feature, which is the same to our method. For outdoor navigation instructions, the CCC~\cite{wang2022counterfactual} speaker is based on an navigation agent model which is not available on UrbanWalk, so it is excluded from UrbanWalk comparison. We also report the results from our implemented ASSISTER~\cite{huang2022assister} model.


\subsection{Main Results}
\label{sec:main_results}

The results on R2R dataset are summarized in Tab.~\!\ref{table:r2r_results}. As shown in the table, the proposed \textsc{Kefa} speaker outperforms previous methods under the SPICE, SPICE-D, Bleu-1, Bleu-4, Meteor and Rouge metrics both on \texttt{val} \texttt{seen} and \texttt{val} \texttt{unseen} splits. For the most important unseen environment, \textsc{Kefa} speaker has an advantage of +1.6 SPICE and +1.3 SPICE-D over the previous state-the-art methods~\cite{wang2022counterfactual,tan2019learning}. For the outdoor navigation part, the results on UrbanWalk Dataset are given in Tab.~\!\ref{table:results_urbanwalk}. The \textsc{Kefa} speaker also achieves better performances under all metrics. These results verify that the proposed method has a strong generation capability on both indoor and outdoor scenes.

\subsection{Ablation Study}
\label{sec:ablation}

In ablation study, we compare the full model with several ablative designs. We test the ablative models on R2R dataset. The results are analyzed as follows:

\noindent \textbf{Baseline.} Our baseline is the speaker-follower~\cite{fried2018speaker} model, which is a widely-used baseline in previous works~\cite{agarwal2019visual,wang2022counterfactual}. 

\noindent \textbf{Teacher-forcing.} We replace the original student-forcing training scheme in the baseline with the teacher-forcing scheme. The difference is: in student-forcing, the instruction decoder produces the next word based its own predicted word in the last step, and in teacher-forcing the ground-truth word is used. We see that the teacher-forcing \textcircled{\raisebox{-0.75pt}{2}} improves baseline \textcircled{\raisebox{-0.75pt}{1}} to 21.3 SPICE \& 26.8 SPICE-D.

\noindent \textbf{Knowledge Refinement Module.} We conduct experiments on removing the Knowledge Refinement Module and replace the fact summary module with two alternative designs: (1) GRU. The convolutional module is replaced with the GRU~\cite{chung2014empirical} to encode the relation type and relation entity of a common-sense fact, and the average pooling is kept; (2) Attention. In this design, we predict a key and value based on concatenate embedding of type and entity, and predict a query based on the object embedding. Comparing \textcircled{\raisebox{-0.75pt}{3}}, \textcircled{\raisebox{-0.75pt}{4}} and \textcircled{\raisebox{-0.75pt}{5}}, the convolutional knowledge module boost the SPICE from 21.3 to 22.9 and SPICE-D from 28.2 to 29.0, and is more effective than the two alternative designs.

\noindent \textbf{Adaptive Temporal Alignment.} We do the ablation study by first removing both the Attention Coverage Loss and Contrastive Loss, and then remove them separately. As shown in the table, the Attention Coverage Loss \textcircled{\raisebox{-0.75pt}{6}} improves model \textcircled{\raisebox{-0.75pt}{2}} with a +1.3 SPICE and +2.0 SPICE-D, which works slightly better than the Contrastive Loss \textcircled{\raisebox{-0.75pt}{7}}. When these two losses are use jointly in \textcircled{\raisebox{-0.75pt}{8}}, the performance is boosted to 22.7 SPICE and 29.0 SPICE-D.

\begin{table*}[!bth]
	\centering
			\resizebox{0.95\textwidth}{!}{
			\setlength\tabcolsep{3pt}
			\renewcommand\arraystretch{1.0}
	\begin{tabular}{|c||c|c|c|c|c|ccccccc|}
	\hline
	~ & & & & & &\multicolumn{7}{c|}{R2R \texttt{val} \texttt{unseen}}\\
	\cline{7-13}\cline{7-13}
	\multicolumn{1}{|c||}{\multirow{-2}{*}{\#}} & \multirow{-2}{*}{\shortstack{Teacher \\ Forcing}} & \multirow{-2}{*}{\shortstack{Knowledge \\ Enhance}} & \multirow{-2}{*}{\shortstack{Summary \\ Module}} & \multirow{-2}{*}{\shortstack{Attention \\ Loss}} & \multirow{-2}{*}{\shortstack{Contrast \\ Loss}}
	&\textbf{\texttt{SPICE}}\!~$\uparrow$ &\textbf{\texttt{SPICE-D}}\!~$\uparrow$ &\texttt{Bleu-1}\!~$\uparrow$ &\texttt{Bleu-4}\!~$\uparrow$ &\texttt{CIDEr}\!~$\uparrow$ &\texttt{Meteor}\!~$\uparrow$ &\texttt{Rouge}\!~$\uparrow$ \\
	\hline
	\hline
	\textcircled{\raisebox{-0.75pt}{1}}    &\xmark &\xmark &\xmark &\xmark &\xmark &18.9 &25.1 &68.2 &26.3 &37.9 &21.7 &48.0\\
        \textcircled{\raisebox{-0.75pt}{2}}    &\cmark &  &  &  &  &21.3 &26.8 &71.7 &28.0 &42.7 &23.1 &49.8\\
        \textcircled{\raisebox{-0.75pt}{3}}    &\cmark &\cmark &\text{GRU} &  &  &22.5 &28.2 &73.1 &27.5 &42.1 &24.1 &49.9\\
        \textcircled{\raisebox{-0.75pt}{4}}    &\cmark &\cmark &\text{Att} &  &  &22.4 &28.8 &73.2 &28.4 &43.6 &24.2 &50.1\\
        \textcircled{\raisebox{-0.75pt}{5}}    &\cmark &\cmark &\text{Conv} &  &  &22.9 &29.0 &73.5 &28.3 &43.2 &24.4 &50.4\\
        \textcircled{\raisebox{-0.75pt}{6}}    &\cmark &  &  & \cmark  &  &22.6 &28.8 &73.5 &28.2 &44.8 &24.2 &50.4\\
        \textcircled{\raisebox{-0.75pt}{7}}    &\cmark &  &  &   & \cmark  &22.4 &28.6 &73.6 &28.1 &41.9 &24.1 &50.0\\
        \textcircled{\raisebox{-0.75pt}{8}}    &\cmark &  &  & \cmark & \cmark  &22.7 &29.0 &72.4 &27.3 &42.4 &24.1 &50.3\\
        \textcircled{\raisebox{-0.75pt}{9}}    &\cmark &\cmark &\text{Conv}  & \cmark  & \cmark  &23.4 &29.3 &73.8 &28.3 &42.7 &24.4 &50.3\\
\hline
	\end{tabular}
	}
	\captionsetup{font=small}
		\caption{\small{\textbf{Ablation Study} Results (\S\ref{sec:ablation}) on R2R~\!\cite{anderson2018vision}.}}
		\label{table:ablation}
\end{table*}

\subsection{SPICE-D Analysis}
\label{sec:spice_d_ana}

The SPICE-D metric is designed to incorporate the correctness of directional expressions into the existing language metric. In this section, we give a further analyze on the SPICE-D metric in order to show the correlations between metrics and human judgements.
We randomly select 150 instructions with at least two directional phrases from the R2R \texttt{val} \texttt{unseen} dataset, and a human annotator evaluates the generated instructions according to the GT by scoring each instruction from 1 to 7. Then  we compute the Pearson correlation coefficient~\cite{schober2018correlation} between each metric and the human evaluation scores. The results are reported in Tab.~\!\ref{table:correl}.
As shown in the table, the original SPICE metric is more correlated (0.543) than the other standard language metrics like CIDEr~\cite{vedantam2015cider} (0.398) and Meteor~\cite{banerjee2005meteor} (0.414). The proposed SPICE-D metric improves the correlation of SPICE and has the highest correlation (0.590) to the human judgement scores among all the metrics.

\begin{table}[h]
	\centering
			\resizebox{0.275\textwidth}{!}{
			\setlength\tabcolsep{3pt}
			\renewcommand\arraystretch{1.0}
	\begin{tabular}{|c||c|c|c|}
	\hline
        Metric    &Pearson Correl.\\
        \hline
        \hline
	Bleu-1~\cite{papineni2002bleu}   &0.367 \\
        Bleu-4~\cite{papineni2002bleu}   &0.252 \\
        CIDEr~\cite{vedantam2015cider}   &0.398 \\
        Meteor~\cite{banerjee2005meteor}   &0.414 \\
        Rouge~\cite{lin2004rouge}   &0.380 \\
        SPICE~\cite{anderson2016spice}   &0.543 \\
\hline
        SPICE-D   &\textbf{0.590} \\
\hline
	\end{tabular}
	}
	\captionsetup{font=small}
		\caption{\small{Correlation Analysis (\S\ref{sec:spice_d_ana})}}
		\label{table:correl}
	\vspace*{-6pt}
\end{table}

\subsection{Qualitative Case Study}
\label{sec:case_study}



\begin{figure}[t]
\begin{center}
\includegraphics[width=1.0\linewidth]{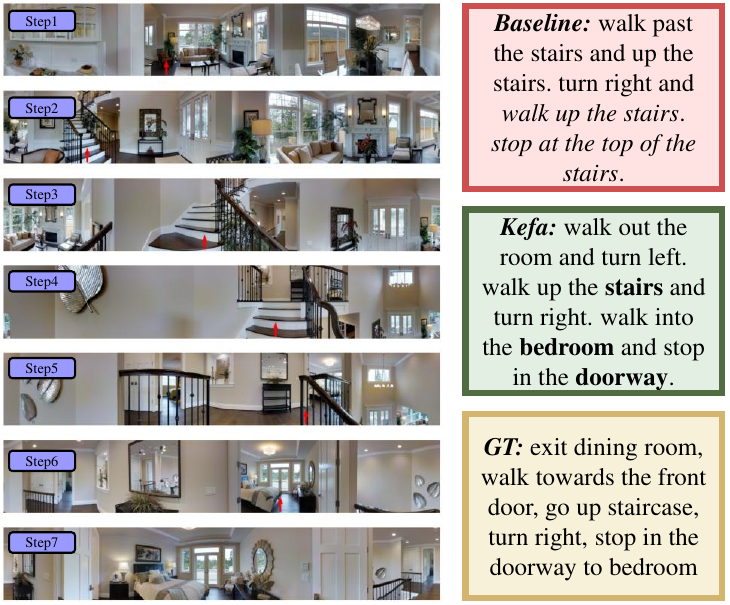}
\end{center}
\vspace{-1em}
\caption{Visualization of the path and instruction generation results (\S\ref{sec:case_study}) on R2R \texttt{val} \texttt{unseen} dataset.}
\vspace{-1em}
\label{fig:path_vis}
\end{figure}

We demonstrate the navigation instruction generation results of a trajectory and make a comparison with the baseline in Fig.~\!\ref{fig:path_vis}. As shown in the figure, the path consists of 7 steps, and the action direction at each step is marked with a red arrow. In this case, the baseline method mentions the \texttt{stairs} repeatedly in three sub-instructions, and fails to describe the stopping position of the trajectory. In contrast, since our \textsc{Kefa} speaker learns the alignment between sub-instruction and sub-trajectories and is aware of the correlation between landmarks \texttt{bed}, \texttt{mirror} and room \texttt{bedroom}, our \textsc{Kefa} speaker generates a high-quality navigation instruction for the path, which contains reasonable directions, rooms and landmarks in the correct order.

\section{Conclusion}
We introduce the \textsc{Kefa} speaker, a high-performance navigation instruction generator. We propose a Knowledge Refinement Module to enhance the vision feature with semantic knowledge facts. We design an Adaptive Temporal Alignment to strengthen the fine-grained correspondence between sub-instructions and sub-trajectories. Furthermore, we introduce a new SPICE-D metric which is aware of direction correctness. The experiments on R2R and UrbanWalk verify that \textsc{Kefa} achieves state-of-the-art instruction generation performance in indoor and outdoor scenarios.

{\small
\bibliographystyle{ieee_fullname}
\bibliography{egbib}
}

\end{document}